\documentclass[11pt,a4paper]{article}
\pdfoutput=1
\usepackage[hyperref]{emnlp-ijcnlp-2019}
\usepackage{times}
\usepackage{latexsym}
\usepackage{multirow}
\usepackage{soul}
\usepackage{color}

\usepackage{graphicx}
\usepackage{caption}
\usepackage{subcaption}

\soulregister\ref{7}
\soulregister\citep{7}
\soulregister\citep{7}
\soulregister\citet{7}
\soulregister\textbf{7}
\soulregister\textit{7}

\usepackage{makecell}
\usepackage{enumitem}
\usepackage{lipsum}
\usepackage{blindtext}
\usepackage{natbib}

\usepackage{url}

\aclfinalcopy 

\begin{document}

\title{Extracting UMLS Concepts from Medical Text Using General and Domain-Specific Deep Learning Models} 

\author{
	Kathleen C. Fraser$^1$\thanks{ \enspace These authors contributed equally.} , Isar Nejadgholi$^{1*}$, Berry De Bruijn$^1$, Muqun Li$^2$,\\ \textbf{Astha LaPlante}$^2$ \textbf{, Khaldoun Zine El Abidine}$^2$
	\\ $^1$ National Research Council Canada, Ottawa, Canada 
	\\ $^2$ Privacy Analytics Inc., Ottawa, Canada
	\\ \tt \footnotesize \{kathleen.fraser, isar.nejadgholi, berry.debruijn\}@nrc-cnrc.gc.ca
	\\ \tt \footnotesize \{rachel.li, astha.agarwal, khaldoun.zineelabidine\}@privacy-analytics.com
}

\maketitle{}

\begin{abstract}


Entity recognition is a critical first step to a number of clinical NLP applications, such as entity linking and relation extraction. We present the first attempt to apply state-of-the-art entity recognition approaches on a newly released dataset, MedMentions. This dataset contains over 4000 biomedical abstracts, annotated for UMLS semantic types. In comparison to existing datasets, MedMentions contains a far greater number of entity types, and thus represents a more challenging but realistic scenario in a real-world setting. We explore a number of relevant dimensions, including the use of contextual versus non-contextual word embeddings, general versus domain-specific unsupervised pre-training, and different deep learning architectures. We contrast our results against the well-known i2b2 2010 entity recognition dataset, and propose a new method to combine general and domain-specific information. While producing a state-of-the-art result for the i2b2 2010 task (F1 = 0.90), our results on MedMentions are significantly lower (F1 = 0.63), suggesting there is still plenty of opportunity for improvement on this new data.

\end{abstract}

\section{Introduction}

Entity recognition is a widely-studied task in clinical NLP, and has been the focus of a number of shared tasks, including the i2b2 2010 Shared Task \cite{Uzuner2011}, SemEval 2014 Task 7 \cite{Pradhan2014}, and SemEval 2015 Task 14 \cite{Elhadad2015}. Most previous work has focused on identifying only a few broad types of entities, such as `problems', `tests', and `treatments' in the i2b2 task, and `diseases' in the SemEval tasks. Even when corpora have been annotated for more entity types, as in the GENIA corpus of biological annotations \cite{Ohta2002}, entity recognition tasks typically focus on only a small subset of those \cite{Kim2004}.

However, in some downstream applications it would be useful to identify \textit{all} terms in a document which exist as concepts in the Unified Medical Language System (UMLS) Metathesaurus \cite{Bodenreider2004}. This resource comprises a much wider range of biomedical entity types than has previously been considered in clinical entity recognition. Additionally, the UMLS Metathesaurus defines important relationships between entity types (and the lower-level concepts associated with them) via its Semantic Network. Therefore, extracting and labelling entities with respect to their UMLS semantic type, rather than more generic types such as `problem' or `test', can be an important first step in many practical clinical NLP applications.

In this work, we present the first attempt to apply existing clinical entity recognition approaches to a new dataset called \textit{MedMentions}, which is annotated for all UMLS semantic types \cite{Mohan2019}. We compare the effectiveness of these approaches with reference to a well-known baseline dataset (i2b2 2010) and analyze the errors that occur when applying such techniques to new problems. On the basis of this error analysis, we propose an adaptation to the BERT architecture to better combine the general and clinical knowledge learned in the pre-training phase, and show that this improves over the more basic approaches.

\section{\label{sec:background}Background}

Early successes in clinical/biomedical entity extraction employed approaches such as conditional random fields \cite{Jonnalagadda2012, Fu2014, Boag2015} and semi-Markov models \cite{deBruijn2011}, requiring numerous engineered features. In recent years, such approaches have been surpassed in performance by deep learning models \cite{Habibi2017}. However, there is a wide range of variation possible within this set of techniques. We briefly discuss some of the parameters of interest in the following sections. 

\subsection{General vs. Domain-Specific Word Embeddings} 
Since words may have one dominant meaning in common use, and a different meaning in the medical domain, some work has explored whether word embeddings trained on medical text (e.g.\@ clinical notes, medical journal articles) are more effective in medical entity recognition than those trained on general text sources (e.g.\@ news, Wikipedia). 

\citet{Roberts2016} examined the effect of training word embeddings on different corpora for the task of entity extraction on the i2b2 2010 dataset. He compared six corpora: the i2b2 dataset itself, the clinical notes available in the MIMIC database \cite{Johnson2016}, MEDLINE article abstracts, WebMD forum posts, and generic text corpora from Wikipedia and Gigaword. It was found that the best F1 score was obtained by training on the MIMIC corpus, and that combining corpora also led to strong results. \citet{Si2019} also compared training embeddings on MIMIC data versus general domain data, and similarly found that pre-training on the MIMIC data led to better performance on both the i2b2 2010 and SemEval tasks. \citet{Alsentzer2019} trained embeddings only on the discharge summaries from MIMIC, and reported a marginal improvement on the i2b2 2010 task over using the entire MIMIC corpus. \citet{Peng2019} found that pre-training a BERT model on PubMed abstracts led to better performance for biomedical entity extraction, while pre-training on a combination of PubMed abstracts and MIMIC notes led to better performance when extracting entities from patient records. 

\subsection{Contextual vs. Non-Contextual Word Embeddings} For many years, word embeddings were non-contextual; that is, a word would have the same embedding regardless of the context in which it occurred. Popular word embeddings of this type include GloVe \cite{Pennington2014}, word2vec \cite{Mikolov2013}, and FastText \cite{Bojanowski2017}. \citet{Peters2018} popularized the idea of contextualized word embeddings, which allowed the same word to have a different representation, depending on the context. The character-based ELMo word embeddings introduced by \citet{Peters2018} can be used just as the non-contextual word embeddings were. \citet{sheikhshabbafghi2018domain} trained ELMo word embeddings on a dataset of biomedical papers and achieved a new state of the art in gene mention detection on the BioCreative II gene mention shared task. This work showed that domain-specific contextual embeddings improve various types of biomedical named entities recognition. Later in 2018, BERT embeddings were also introduced \cite{Devlin2019}. The BERT architecture improved over ELMo by using a different training objective to better take into account both left and right contexts of a word, and made it possible to make use of the entire pre-trained network in the downstream task, rather than simply extracting the embedding vectors. 

\citet{Si2019} compared word2vec, GloVe, FastText, ELMo, and BERT embeddings on the i2b2 2010 dataset. When using the pre-trained vectors (trained on general-domain corpora), BERT-large performed the best and word2vec performed the worst, but there was no clear advantage to the contextualized embeddings (e.g. GloVe performed better than ELMo). When the embeddings were pre-trained on MIMIC data, the contextualized embeddings did perform appreciably better than the non-contextualized embeddings. 

\subsection{Classifier Architecture} 
Much of the recent work on medical entity extraction has made use of the Long Short-Term Memory (LSTM) architecture \cite{Hochreiter1997}, with some variations and modifications: (1) most work uses a bi-directional LSTM (bi-LSTM), so the prediction for any word in the sequence can take into account information from both the left and right contexts, (2) some work additionally feeds the output of the bi-LSTM layer into a CRF classifier \cite{Huang2015,Chalapathy2016,lample2016neural,Habibi2017,tourille2018evaluation}, to predict the most likely sequence of labels, rather than just the most likely label for each word independently, and (3) some models incorporate additional information (e.g. character embeddings, or traditionally engineered features) at various points in the model \cite{unanue2017recurrent}. 

In contrast, the BERT model makes use of the Transformer architecture, an attention-based method for sequence-to-sequence modelling \cite{Vaswani2017}. Once the model has been pre-trained, in the entity extraction stage it is only necessary to add a simple classification layer on the output. 
However, others have also experimented with 
feeding the output of the BERT model to a bi-LSTM \cite{Si2019}. 

\begin{table*}[htb!]
\small
\centering
\begin{tabular}{l | >{\raggedleft\arraybackslash}p{1.5cm}>{\raggedleft\arraybackslash}p{1.5cm}| >{\raggedleft\arraybackslash}p{1.5cm}>{\raggedleft\arraybackslash}p{1.5cm} | >{\raggedleft\arraybackslash}p{1.5cm}>{\raggedleft\arraybackslash}p{1.5cm}  }
\Xhline{3\arrayrulewidth}
 & \multicolumn{2}{c|}{i2b2 2010} & \multicolumn{2}{c|}{MedMentions (full)} & \multicolumn{2}{c}{MedMentions (st21pv)} \\
 \hline 
& Train & Test & Train & Test & Train & Test \\ 
\Xhline{3\arrayrulewidth}
\# entity types & 3         & 3         & 126       & 123       & 21        & 21 \\ 
\# documents    & 170       & 256       & 3513      & 879       & 3513      & 879  \\ 
\# tokens       & 149,743   & 267,837   & 936,247   & 234,910   & 936,247   & 234,910 \\ 
\# entities     & 16,520    & 31,161    & 281,719   & 70,305    & 162,908   & 40,101 \\ 
\Xhline{3\arrayrulewidth}

\end{tabular}
\caption{\label{tab:datasets}Properties of the datasets. For MedMentions, we combine the training and validation sets into `Train'.}
\end{table*}

\section{Methods}

\subsection{Data}

We consider three datasets in this study: the i2b2 2010 dataset, the `full' MedMentions dataset, and the `st21pv' MedMentions dataset. Details are shown in Table~\ref{tab:datasets}.

The i2b2 2010 corpus\footnote{\href{http://www.i2b2.org/NLP/DataSets}{www.i2b2.org/NLP/DataSets}} consists of de-identified clinical notes (discharge summaries), annotated for three entity types: problems, tests, and treatments \cite{Uzuner2011}. The original shared task also included subtasks on assertion classification and relation extraction, but we focus here on entity extraction. In the original data release for the shared task, the training set contained 394 documents; however, the current release of the dataset contains only 170 training documents. Therefore, it is unfortunately not possible to directly compare results across the two versions of the dataset. However, a majority of the recent works are using the current release of the dataset \cite{Zhu2018,bhatia2019joint,chalapathy2016bidirectional}. 

The MedMentions corpus was released earlier this year\footnote{\href{http://github.com/chanzuckerberg/MedMentions}{github.com/chanzuckerberg/MedMentions}}, and contains 4,392 abstracts from PubMed, annotated for concepts and semantic types from UMLS (2017AA release). UMLS \textit{concepts} are fine-grained biomedical terms, with approximately 3.2 million unique concepts contained in the metathesaurus \cite{Mohan2019}. Each concept is linked to a higher-level \textit{semantic type}, such as `Disease or syndrome', `Cell component', or `Clinical attribute'. In this work we focus on identifying the semantic type for each extracted text span, leaving the concept linking/normalization for future work. The creators of the dataset have defined an official 60\%-20\%-20\% partitioning of the corpus into training, development, and test sets. 

There are 127 semantic types in UMLS. Of these, there is only one (`Carbohydrate sequence') which never appears in the full MedMentions dataset. Approximately 8\% of the concepts in UMLS can be linked to more than one semantic type \cite{Mohan2019}; in such cases the dataset contains a comma-separated list of all these type IDs corresponding to alphabetical order of semantic types. Where a text span has been labelled with more than one label, we select only the first one. As a result of this, there is one other type (`Enzyme') which appears in MedMentions, but only doubly-labelled with `Amino acid, peptide, or protein', and thus does not occur in our singly-labelled training or test data. Finally, there is an extra class (`UnknownType'), for a total of 126 semantic types or classes in the `full' training data. Of these, there are three (`Amphibian', `Drug delivery device', and `Vitamin') which never occur in the test data. 

The full MedMentions dataset suffers from high class imbalance (e.g.\@ there are 31,485 mentions for the semantic type `Qualitative concept' and only two mentions for `Fully formed anatomical structure'). Furthermore, many of the semantic types are not particularly useful in downstream clinical NLP tasks, either due to being too broad or too specialized. As a result, the creators of the MedMentions dataset also released an alternate version called `st21pv', which stands for `21 semantic types from preferred vocabularies'. The details of how this subset was constructed are given by \citet{Mohan2019}, but essentially it contains only 21 semantic types, from specific vocabularies most relevant to biomedical researchers. The raw abstracts, and partitions into training, development, and test sets are the same as in the full dataset -- only the set of annotations differs.

The i2b2 and MedMentions datasets differ across a number of important dimensions: the discharge summaries in the i2b2 dataset tend to be hastily written or dictated, with short, incomplete sentences and numerous acronyms and abbreviations, compared to the academic writing style of the MedMentions abstracts. The discharge summaries also tend to be longer, averaging approximately 980 tokens per document, compared to 267 tokens per document in MedMentions. The semantic content of the documents is also different, with the discharge summaries focused exclusively on a single patient and their history, disease progression, treatment, and outcomes, while the MedMentions abstracts typically summarize the results of a scientific study, covering a wide range of biomedical topics. Finally, there are clearly far more entity types in MedMentions than in i2b2, and greater imbalance between the different classes. Therefore, there is no guarantee that methods which perform well on the i2b2 data will also be effective on the MedMentions dataset.

\subsection{Entity Recognition}

Based on our review of the literature (Section~\ref{sec:background}) we experimented with two basic architectures: bi-LSTM+CRF (with pre-trained contextual and non-contextual word embeddings as input), and BERT (with a simple linear classification layer and a bi-LSTM classification layer). The details of these classifiers and their training are described below. 


\subsubsection{Text pre-processing}

For the bi-LSTM+CRF models, input text retained casing information, but all numerical tokens were normalized to a single NUM token. 

BERT uses WordPiece tokenization \cite{Wu2016},  which breaks longer words into frequently occurring sub-word units to improve handling of rare words and morphological variation. This requires additional pre- and post-processing for the entity recognition task, since the data is labelled at the word level. Following the recommendation of \citet{Devlin2019}, we first re-tokenize the text using the WordPiece tokenizer, assign the given label to the first piece of each word, and assign any subsequent pieces a padding label.

In all cases, we convert the text and labels to CoNLL IOB format for input to the classifiers. 

\subsubsection{bi-LSTM+CRF}

We use a standard bi-LSTM+CRF architecture (e.g., see \cite{huang2015bidirectional}), implemented in PyTorch. The bi-LSTM component has 2 bi-directional layers with hidden size of 1536 nodes.  The 100-dimensional character embeddings are learned through the training process and concatenated with pre-trained GloVe embeddings \cite{Pennington2014} as proposed by \citet{Chalapathy2016}.  We compare the performance of general GloVe embeddings, trained on Wikipedia and Gigaword, and clinical GloVe embeddings, trained on the MIMIC-III corpus \cite{Johnson2016}. In both cases the GloVe embeddings have 300 dimensions. For pre-training on MIMIC, we used a minimum frequency cut-off of 5, and a window size of 15.

We also experimented with contextual ELMo embeddings \cite{Peters2018} and the bi-LSTM+CRF architecture, comparing general ELMo embeddings\footnote{\href{https://github.com/allenai/allennlp/blob/master/tutorials/how_to/elmo.md}{github.com/allenai/allennlp/blob/master/tutorials/how\_to/elmo.md}} with clinical ELMo embeddings.\footnote{\href{https://github.com/noc-lab/clinical_concept_extraction}{github.com/noc-lab/clinical\_concept\_extraction}} The clinical ELMo embeddings were released by \citet{Zhu2018} and trained on Wikipedia pages whose titles are medical concepts in the SNOMED-CT vocabulary, as well as MIMIC-III.

The bi-LSTM+CRF models were trained using the Adam optimizer with a learning rate of 0.001 and a batch size of 32 for 10 epochs. 

\subsubsection{BERT}

The BERT (Bidirectional Encoder Representations from Transformers) model is described by \citet{Devlin2019} and proposes to address some of the limitations observed in LSTM models. 
In our experiments, we use the BERT-base architecture, which has 12-layers, hidden size 768, and 12 self-attention heads. To perform the entity recognition, we added a linear layer and a softmax layer on top of the last BERT layer to determine the most probable label for each token. While this is the approach taken by 
\citet{Alsentzer2019}, others suggest using a more complex classification model in conjunction with BERT \cite{Si2019}, and so we also experiment with a bi-LSTM layer with input and output size of $4\times768$ 
on top of the concatenation of the last four layers of BERT.  

We consider four pre-trained BERT models:
\begin{itemize}[noitemsep]
    \item \textbf{BERT-base}\footnote{\href{https://github.com/google-research/bert}{github.com/google-research/bert}} General domain BERT model released by Google, pre-trained on Wikipedia and BookCorpus \cite{Devlin2019}.
    \item \textbf{bioBERT (v1.1)}\footnote{\href{https://github.com/dmis-lab/biobert}{github.com/dmis-lab/biobert}} The bioBERT model is initialized with BERT-base, and then further pre-trained on biomedical abstracts from PubMed \cite{Lee2019}. 
    \item \textbf{clinicalBERT}\footnote{\href{https://github.com/EmilyAlsentzer/clinicalBERT}{github.com/EmilyAlsentzer/clinicalBERT}}  The clinicalBERT model is initialized with bioBERT, and then further pre-trained on clinical notes from the MIMIC corpus \cite{Alsentzer2019}.
    \item \textbf{NCBI BERT}\footnote{\href{https://github.com/ncbi-nlp/NCBI_BERT}{github.com/ncbi-nlp/NCBI\_BERT}} The NCBI BERT model is initialized with BERT-base, and then further pre-trained on PubMed abstracts and MIMIC notes \cite{Peng2019}.
\end{itemize}

In the fine-tuning stage, we generally follow the recommendations in \citet{Devlin2019}, and use an Adam optimizer with $\beta_1 = 0.9$, $\beta_2 = 0.999$, L2 weight decay of 0.01, and a dropout probability of 0.1 on all layers. We use a learning rate warmup over the first 10\% of steps, and linear decay of the learning rate thereafter. Before training the final models, we conducted a series of hyper-parameter optimization experiments using 10-fold cross-validation on the training set. In this optimization stage we considered combinations of batch sizes in $\{16, 32\}$, learning rates in 
$\{0.00002, 0.00003, 0.00005, 0.0001\}$, and number of training epochs in $\{1 ... 10\}$. We also determined that the uncased BERT-base model led to marginally better results, and so we use that in our final evaluation (bioBERT and clinicalBERT are cased, while NCBI BERT is uncased). For the BERT+bi-LSTM model, we also experimented with training only the bi-LSTM component and not fine-tuning the pre-trained layers, but found that fine-tuning led to better results in development.


\subsection{Evaluation}

To evaluate the systems, we use micro-averaged strict precision, recall, and F-score. This means that for any given recognized entity, it is only counted as a true positive if \textit{both} the span and the label match exactly with the gold standard annotation. Note also that these metrics are computed on the entity-level, not the token level. For example, given the following gold and predicted label sequences:

\texttt{GOLD: O O B-prob I-prob I-prob}

\texttt{PRED: O O B-prob I-prob O}

\noindent A token-level evaluation would identify two true positives, but a strict entity-level evaluation identifies zero true positives.

\section{Results}

Table \ref{tab:results} shows the results of the entity recognition experiments for each model and dataset. 

\begin{table*}[tbh!]
\small
\centering
\begin{tabular}{p{3.1cm}|  p{2.2cm} | rrr | rrr | rrr   }
\Xhline{3\arrayrulewidth}
 & & \multicolumn{3}{c|}{\textbf{i2b2 2010}} & \multicolumn{3}{c}{\textbf{MedMentions}} & \multicolumn{3}{c}{\textbf{MedMentions}}\\
& & \multicolumn{3}{c|}{} & \multicolumn{3}{c}{\textbf{(full)}} & \multicolumn{3}{c}{\textbf{(st21pv)}}\\
  \hline
 \textbf{Model} & \textbf{Domain} & P & R & F1 & P & R & F1 & P & R & F1  \\
\Xhline{3\arrayrulewidth}
Glove + bi-LSTM+CRF   & general           &0.81 &0.76 &0.79 &0.54 &0.51 &0.52 &0.60 &0.50 &0.54 \\
\hline 
Glove + bi-LSTM+CRF & clinical        &0.83 &0.77 &0.80 &0.45 &0.37 &0.41 &0.59 &0.46 & 0.52\\
\hline

\Xhline{3\arrayrulewidth}
ELMo + bi-LSTM+CRF   & general           &0.80 &0.80 & 0.80&0.43 & 0.45& 0.44&0.54 &0.50 &0.52 \\
\hline 
ELMo + bi-LSTM+CRF   & clinical          &0.86 &0.86 &0.86 & 0.47& 0.47 &0.47 & 0.58& 0.53&0.56 \\
\Xhline{3\arrayrulewidth}

BERT-base + linear & general           &0.85 &0.87 &0.86 &0.51 &0.55 &0.53 & 0.58 & 0.61 & 0.59 \\
\hline 
bioBERT + linear & biomed        &0.86 &0.88 & 0.87&0.53 &0.57 &0.55 & 0.61 & 0.64 & 0.62 \\
\hline 
clinicalBERT + linear & biomed + clinical          &0.87 &0.88 & 0.88& 0.51 & 0.56 & 0.53 &0.59 &0.62 &0.61 \\
\hline 
NCBI BERT + linear & biomed + clinical  & \textbf{0.88} &\textbf{0.90} & \textbf{0.89}& 0.51 & 0.56 & 0.53 &0.59 &0.61 &0.60 \\
\Xhline{3\arrayrulewidth}
bioBERT + bi-LSTM       & biomed &0.86&0.88&0.87&\textbf{0.53} & \textbf{0.58} & \textbf{0.56} &\textbf{0.61} &\textbf{0.66} &\textbf{0.63} \\
\hline
NCBI BERT + bi-LSTM       & biomed + clinical&\textbf{0.88}&\textbf{0.90}& \textbf{0.89}&0.52 &0.57 & 0.54&0.59 &0.62 & 0.60\\


\Xhline{3\arrayrulewidth}

\end{tabular}
\caption{\label{tab:results} Results of entity recognition for each dataset and model.}
\end{table*}

\subsection{Effect of Contextual vs. Non-Contextual Word Embeddings}
If we first consider the bi-LSTM+CRF results for the i2b2 dataset, we observe that the contextual ELMo embeddings lead to better results than the non-contextual GloVe embeddings, and in both cases, better results are obtained by pre-training the embeddings on domain-specific text. For MedMentions, however, for both versions of the dataset we observe that the general-domain GloVe embeddings outperform the clinical GloVe embeddings, but the clinical ELMo embeddings outperform the general ELMo embeddings. \citet{Si2019} also observed a greater benefit to using contextual embeddings when pre-training on domain-specific corpora. Here, this may  be due in part to differences between the training corpora; for example, clinical GloVe was trained only on MIMIC notes, while clinical ELMo was trained on a combination of MIMIC notes and Wikipedia articles about medical concepts, which may be more similar to the biomedical abstracts contained in MedMentions.

The BERT models offer a substantial improvement in F1 over the models based on Glove or ELMo embeddings for each of the three datasets. For the i2b2 dataset, the best results are obtained using clinicalBERT (F1 = 0.88) and NCBI BERT (F1 = 0.89), each of which involved pre-training on clinical notes from MIMIC. This demonstrates the importance of pre-training on documents which are similar in nature to those seen in the labelled dataset. Consistent with this, the bioBERT model (pre-trained on biomedical abstracts) leads to the best result on both MedMentions datasets. 

\subsection{Effect of Classifier Structure}

Finally, comparing the effectiveness of a simple linear model versus a bi-LSTM model as the top BERT layer, we observe that this change makes no difference on the i2b2 dataset, but leads to the best result for both versions of MedMentions. It may be that the much larger training set in MedMentions is better able to effectively train the more complex classifier, and respectively that the greater complexity is necessary to properly model the large number of classes in MedMentions. 

\subsection{MedMentions vs i2b2 dataset}

Comparing across datasets, it is clear that performance is worse on the MedMentions data than the i2b2 data. One obvious reason for this is that the MedMentions datasets have many more entity types, or classes, than the i2b2 dataset. This means that the number of examples per class seen in the training data is much lower (in some cases, only a handful), and also that the classes tend to be more easily confused. The ambiguity between classes arises at the level of the training data where, depending on context, the same text span will be associated with different labels (e.g. \textit{neurocognitive function} is sometimes labelled as a `biologic function' and sometimes as a `mental process'; \textit{PSA levels} is labelled as both a `laboratory procedure' and a `laboratory or test result'; and an even more highly-ambiguous term such as \textit{probe} is variously labelled as a `medical device', `indicator reagent or diagnostic aid', `nucleic acid nucleoside or nucleotide', `functional concept', `diagnostic procedure', `research activity', and `chemical viewed functionally'). Thus in MedMentions, the context becomes extremely important, and fine-grained distinctions between entity types must be  learned.


\section{Error Analysis}

In the following section, we examine the errors made by the entity recognition systems from two different perspectives: first, we compare generally the types of errors made on the two datasets; then, we consider the role of general versus domain-specific pre-training and examine some of the specific errors that occur in each case.

\subsection{Types of Errors by Dataset}


\begin{figure*}
\centering
\begin{subfigure}{.5\textwidth}
  \centering
  \includegraphics[trim={0.1cm -0.5cm 0.1cm 0.5cm},clip,width=1\linewidth]{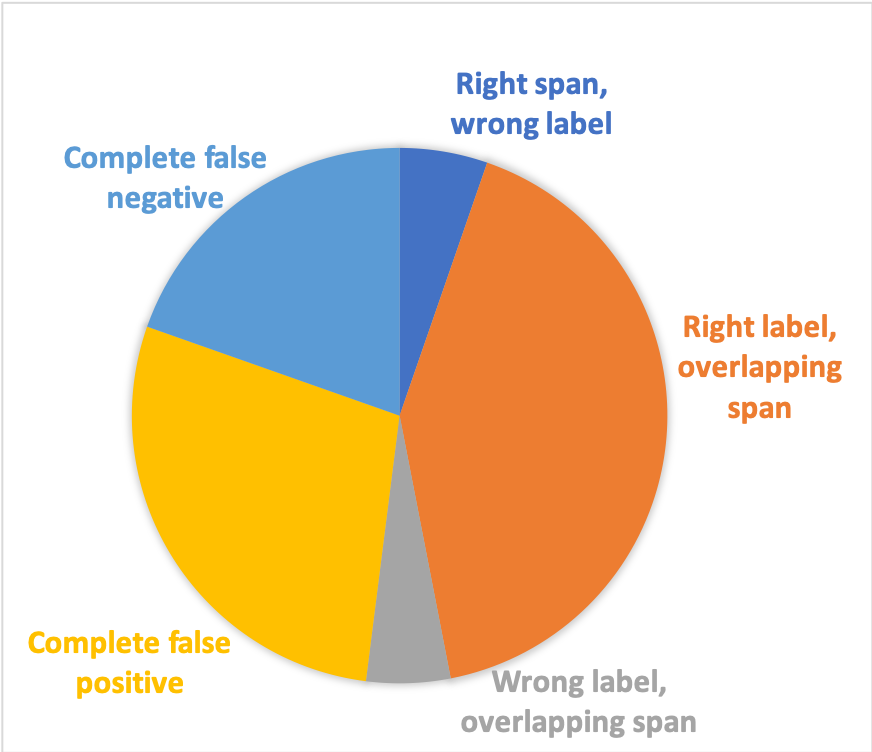}
  \caption{i2b2 2010 (NCBI BERT + linear)}
  \label{fig:errors_i2b2}
\end{subfigure}%
\begin{subfigure}{.5\textwidth}
  \centering
  \includegraphics[trim={0.1cm 0.1cm 0.1cm 0.1cm},clip,width=1\linewidth]{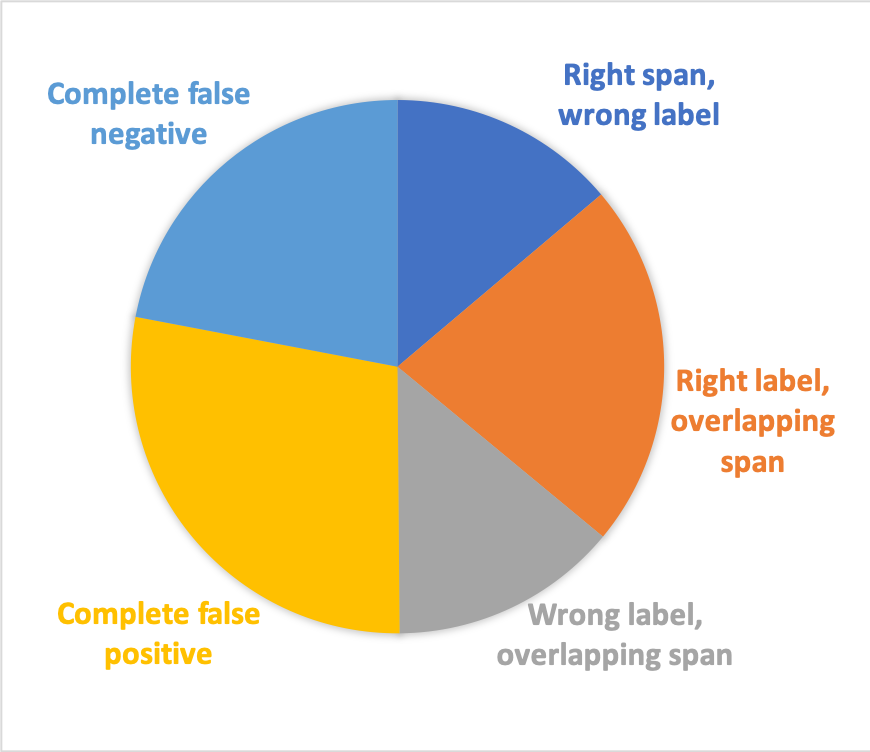}
  \caption{MedMentions-st21pv (bioBERT + linear)}
  \label{fig:errors_MedMentions}
\end{subfigure}
\caption{Types of errors made on the i2b2 and MedMentions-st21pv datasets.}
\label{fig:errors}
\end{figure*}

Depending on the downstream application of entity recognition, different types of errors may be associated with different costs. For example, if a company is using this model in practice, the cost associated with having human annotators adjust label boundaries may be different from the cost associated with having them search for entities which have been missed altogether. Our evaluation metrics, however, do not reflect the differences among various types of errors. To further investigate the nature of the errors being made by the system, we investigated three specific types of `partial errors'. These cases are counted as \textit{false} in calculating the evaluation metrics, but the model actually gets at least part of the information correct:

\begin{itemize}[noitemsep]

 \item \textbf{Right span, wrong label:} the text span associated with an entity is correctly identified, but assigned the wrong label.

\item \textbf{Right label, overlapping span:} the entity is correctly labelled, but the text span associated with the entity is not exactly the same as that indicated in the gold transcripts. 

\item \textbf{Wrong label, overlapping span:} the entity overlaps with one of the gold entities, but is assigned the wrong label. 

\end{itemize}

In addition to these categories, errors can be complete false positives (model extracts an entity which does not overlap at all with any gold entities), or complete false negatives (model completely misses an entity from the gold transcripts). 

In the i2b2 dataset, over half of the errors belong to one of the `partial' error types (Figure~\ref{fig:errors_i2b2}), with the biggest error category overall being \textbf{right label, overlapping span}. In many cases, the model identifies the main noun phrase associated with the entity, but not adjectives or prepositional phrases that further describe the entity: e.g., \textit{patient feeling weaker than usual}, where the model labels \textit{weaker} as the problem but the gold entity is \textit{weaker than usual}, or \textit{her glucose remained somewhat low}, where the model labels \textit{low} as the problem, but the gold entity is \textit{somewhat low}. One of the reasons that might lead to this kind of error is the inconsistency that exists in the human annotations. The task of identifying the span of entities can be very subjective and there is always some level of disagreement among human annotators. Also, sometimes annotation guidelines are interpreted differently by various annotators. As a result of this inconsistency, in the training stage, the model sees examples that the spans of the same entities has been labeled differently. This type of error may occur more frequently in the i2b2 dataset, due to the difficulty in annotating hurriedly written notes compared to academically written abstracts. 

For MedMentions-st21pv, we again observe that roughly half the errors are `partial' errors, but with a sizable increase in errors that involve the wrong label, with either a correct or overlapping span (Figure~\ref{fig:errors_MedMentions}). As discussed previously, the increase in \textbf{right span, wrong label} is likely due to the higher ambiguity between entity types in this dataset. In many cases, errors of the type \textbf{wrong label, overlapping span} appear to be due to 
the inability of the annotation scheme to handle overlapping entities. 
For example,  in \textit{co-expression analysis}, where the model labels \textit{co-expression} as a `biologic function', rather than extracting the entire phrase as a `research activity', or in \textit{this region of the brain}, where the model simply labels \textit{brain} as an `anatomical structure' rather than \textit{region of the brain} as a `spatial concept'.

\subsection{General vs Domain Knowledge}

We then performed an exploratory error analysis to identify and compare the type of errors made by models using \textit{general} and \textit{domain-specific} embeddings. For this analysis, we considered BERT-base and NCBI BERT. BERT-base is pre-trained on Wikipedia and the Google Books database, so it transfers the general knowledge of language to the model. NCBI BERT starts its training from BERT-base, and then continues training on PubMed (biomedical abstracts) and MIMIC (clinical notes).  Generally, embeddings like NCBI BERT are assumed to be more effective since they transfer both general and domain-specific knowledge. We analyze the errors of the two models to test this assumption. For simplicity, we concentrate our analysis on the i2b2 dataset, although we observe similar patterns in MedMentions.

NCBI BERT results in a higher overall F1-score than BERT-base on the i2b2 dataset, and there are 2027 entities that are correctly recognized by NCBI BERT and incorrectly recognized by BERT-base. However, there are also 1209 entities that are \textit{correctly} recognized by BERT-base but \textit{incorrectly} recognized by NCBI BERT. Therefore, it is not the case that NCBI BERT encodes the same knowledge as BERT-base, plus more that it has learned from PubMed and MIMIC; rather, the two systems have different strengths and weaknesses.

Qualitatively, we observed that some entities correctly recognized by NCBI BERT and missed by BERT-base involve common words that have a specialized meaning in medicine, for example in the sentence: \textit{Suck , root , gag , grasp , and morrow were normal .} The BERT-base model does not extract any entities, while the NCBI BERT model recognizes \textit{suck}, \textit{root}, \textit{gag}, and \textit{grasp} as standard tests of infant reflexes. NCBI BERT also appears to be better at recognizing specialized acronyms and abbreviations, particularly when there is very little context, as in  \textit{Brother died 64 / MI}, where only NCBI BERT recognizes \textit{MI} as a problem (myocardial infarction) in this brief family history, or \textit{No CPR / No defib}, where NCBI BERT correctly labels \textit{CPR} and \textit{defib} as treatments, while BERT-base mis-labels them as problems.

\begin{table*}[ht!]
\small
\centering
\begin{tabular}{p{5.5cm}| lll | lll | lll   }
\Xhline{3\arrayrulewidth}
 &  \multicolumn{3}{c|}{\textbf{i2b2 2010}} & \multicolumn{3}{c}{\textbf{MedMentions}} & \multicolumn{3}{c}{\textbf{MedMentions}}\\
&  \multicolumn{3}{c|}{} & \multicolumn{3}{c}{\textbf{(full)}} & \multicolumn{3}{c}{\textbf{(st21pv)}}\\
  \hline
 \textbf{Model}  & P & R & F1 & P & R & F1 & P & R & F1  \\
\Xhline{3\arrayrulewidth}

NCBI BERT concat BERT-base +linear   &\textbf{0.89*} & \textbf{0.90}& \textbf{0.90*}&0.52 &0.57 &0.54 & 0.59 &0.62 &0.61\\ 
\hline 
bioBERT concat BERT-base +linear  &0.85 &0.88 &0.86 & \textbf{0.54*} & \textbf{0.59*}&\textbf{0.56*} & 0.60  &0.65 & 0.62\\
\hline 
NCBI BERT concat BERT-base +bi-LSTM  &\textbf{0.89*} &\textbf{0.90} &\textbf{0.90*} & 0.50 &0.55&0.53 & 0.59  &0.62 & 0.61\\
\hline 
bioBERT concat BERT-base +bi-LSTM  &0.86 &0.87 &0.87 & 0.53 &0.58&0.55 & \textbf{0.63*}  &\textbf{0.65} & \textbf{0.64*}\\
\hline 

\Xhline{3\arrayrulewidth}

\end{tabular}
\caption{\label{tab:concat_results} Results of entity recognition using concatenated BERT models. An asterisk indicates an improvement over the best result from Table~\ref{tab:results}.}
\end{table*}

In cases where BERT-base does better than NCBI BERT, it may be partially due to a better knowledge of well-formed text. We observed several examples where BERT-base appeared to be better at identifying the appropriate qualifiers to attach to a given entity: e.g., in \textit{no interval development of effusion}, BERT-base correctly extracts the entire phrase \textit{internal development of effusion}, while NCBI BERT only extracts \textit{effusion}. Similarly, in \textit{Right ventricular chamber size and free wall motion are normal}, BERT-base extracts \textit{Right ventricular chamber size} as a single entity of the type `test', while NCBI BERT splits it into \textit{Right ventricular} and \textit{size}.

Of course, these observations are purely anecdotal at this point, and will require future work and annotation to fully quantify the nature of the differences between the models. However, given the fact that the two models make different errors, it is at least reasonable to assume that predictions from the two models can be combined in a complementary fashion to improve the overall performance. We explore one possible architecture for doing so in the following section.









\section{Concatenated Model}

As a result of our error analysis, we propose a concatenated BERT model, to better combine the general knowledge from BERT-base and the clinical knowledge from the more specialized BERT models. To build such a model we concatenate the last encoding layer of a domain-specific BERT model with the last encoding layer of the general BERT model and feed this concatenation to a linear or bi-LSTM classification layer. During training we jointly fine-tune both BERT models and the classification layer. We implemented this model with both NCBI BERT and bioBERT models, since they previously led to the optimal results for i2b2 and MedMentions, respectively. NCBI BERT is concatenated with the uncased BERT-base model and bioBERT is concatenated with the cased BERT-base model. 

Results for the concatenated models are given in Table~\ref{tab:concat_results}. For all three datasets, we observe a small improvement over the best performing models in Table~\ref{tab:results}. The best result for i2b2 is achieved by concatenating the NCBI BERT and BERT-base models, with either a linear or bi-LSTM classifier on top. The resulting F1 score of 0.90 beats the previously reported state-of-the-art of 0.89 on the current release of the dataset with 170 training documents \citep{Zhu2018}. For MedMentions, concatenating bioBERT and BERT-base leads to the best results, with MedMentions-full attaining the best result using a linear classifier and MedMentions-st21pv attaining the best result with the bi-LSTM. To our knowledge, there are no prior results reported on entity (i.e.\@ semantic type) extraction on this dataset. 

Regarding the classifier layer, we observe that replacing a linear classifier layer with a bi-LSTM does not improve the results on i2b2 dataset. This is consistent with the results shown in Table \ref{tab:results} and indicates that a simple linear classifier is \textit{enough} to learn the entity recognition task for i2b2 dataset. In the case of the MedMentions dataset, a bi-LSTM classifier improves the F1 score on MedMentions-st21pv but worsens it on MedMentions-full. These results show that there is room for more rigorous investigation about the classifier layer for extracting entities in MedMentions dataset. More complex neural structures with optimized hyperparameters may be needed to improve these results.

Although the improvements that we see by concatenating the models are relatively small, they are consistent across the three datasets. This suggests that explicitly inputting both general and domain-specific information to the entity recognizer, rather than sequentially pre-training on different domains and hoping that the model `remembers' information from each domain, can be a promising direction for future research.

\section{Conclusion}

We have presented the results of a set of medical entity recognition experiments on a new dataset, MedMentions. We contrasted these results with those obtained on the well-studied i2b2 2010 dataset. We explored a number of relevant dimensions, including the use of various embedding models (contextual versus non-contextual, general versus domain-specific, and LSTM versus attention-based) as well as linear versus bi-LSTM classifier layers. We also proposed a new modification to the previous BERT-based named entity recognition architectures, which allows the classifier to incorporate information from both general and domain-specific BERT embeddings. Our results on i2b2 are state-of-the-art, and our results on MedMentions set a benchmark for future work on this new dataset.

As popular public datasets become more and more studied over time, there is a chance that even if individual researchers follow good train-validate-test protocols, we eventually overfit to the datasets as a \textit{community}, since there is so much published information available about what works well to improve performance on the test set. One goal of this work was to explore the gap in performance between a well-known clinical entity recognition dataset and a new, unstudied dataset. The same models and training procedures lead to significantly lower performance on the MedMentions dataset, for a variety of reasons: greater number of entity types, more class ambiguity, higher class imbalance, etc. 
Ultimately, we find that the model which performs best on i2b2 2010 is \textit{not} the model that performs best on MedMentions, and that results on MedMentions can be improved by pre-training on more similar documents (biomedical abstracts), and by using more complex models (BERT + bi-LSTM rather than BERT + linear). 
We hope that other researchers will continue to advance the state-of-the-art on this new dataset.

\section*{Acknowledgement}
We would like to thank Dr. Khaled Emam for his support and insightful feedback. We also thank Ms Lynn Wei for her technical support and her assistance in data collection, cleaning and preparation.  

\bibliography{main}
\bibliographystyle{acl_natbib}

\end{document}